\documentclass[11pt,a4paper]{article}
\usepackage[hyperref]{emnlp-ijcnlp-2019}
\usepackage{times}
\usepackage{latexsym}
\usepackage{amsmath}
\usepackage{amssymb}
\usepackage{url}
\usepackage{graphicx}
\usepackage{adjustbox}

\aclfinalcopy 

\title{Towards Generalizable Neuro-Symbolic Systems for Commonsense Question Answering}

\author{Kaixin Ma$^{\P}$\thanks{Work was done during an internship at Bosch Research.} \quad Jonathan Francis$^{\P\S}$ \quad Quanyang Lu$^{\dag}$ \quad Eric Nyberg$^{\P}$ \quad Alessandro Oltramari$^{\S}$\\
  $^{\P}$Language Technologies Institute, School of Computer Science, Carnegie Mellon University   \\
  $^{\dag}$Department of Mechanical Engineering, College of Engineering, Carnegie Mellon University \\
  $^{\S}$Intelligent IoT, Bosch Research and Technology Center (Pittsburgh, USA) \\
  {\tt \{kaixinm, jmf1, qlv, ehn\}@cs.cmu.edu}\\ {\tt alessandro.oltramari@us.bosch.com} \\}

\date{}
\begin{document}
\maketitle

\begin{abstract}
Non-extractive commonsense QA remains a challenging AI task, as it requires systems to reason about, synthesize, and gather disparate pieces of information, in order to generate responses to queries. Recent approaches on such tasks show increased performance, only when models are either pre-trained with additional information or when domain-specific heuristics are used, without any special consideration regarding the knowledge resource type. In this paper, we perform a survey of recent commonsense QA methods and we provide a systematic analysis of popular knowledge resources and knowledge-integration methods, across benchmarks from multiple commonsense datasets. Our results and analysis show that attention-based injection seems to be a preferable choice for knowledge integration and that the degree of domain overlap, between knowledge bases and datasets, plays a crucial role in determining model success.
\end{abstract}

\section{Introduction}
With the recent success of large pre-trained language models \cite{devlin-etal-2019-bert,radford2019language,yang2019xlnet,DBLP:journals/corr/abs-1907-11692}, model performance has reached or surpassed human-level capability on many previous question-answering (QA) benchmarks \cite{NIPS2015_5945,rajpurkar-etal-2016-squad,DBLP:journals/corr/LaiXLYH17}. However, these benchmarks do not directly challenge model reasoning capability, as they require only marginal use of external knowledge to select the correct answer, i.e., all the evidence required to solve questions in these benchmarks is explicit in the context lexical space. 
Efforts have been made towards building more challenging datasets that, by design, require models to synthesize external commonsense knowledge and leverage more sophisticated reasoning mechanisms \cite{DBLP:journals/corr/abs-1810-12885,ostermann-etal-2018-semeval}, showing that the previous state-of-the-art models often struggle to solve these newer tasks reliably. As a result, commonsense has received a lot of attention in other areas as well, such as natural language inference \cite{zellers-etal-2018-swag,zellers-etal-2019-hellaswag} and visual question answering \cite{DBLP:journals/corr/abs-1811-10830}. 
Despite the importance of commonsense knowledge, however, previous work on QA methods takes a coarse-grained view of commonsense, without considering the subtle differences across the various knowledge types and resources. 
Such differences have been discussed at length in AI by philosophers, computational linguists, cognitive psychologists (see for instance \cite{davis2014representations}): at the high level, we can identify \textit{declarative commonsense}, whose scope encompassess \text{factual knowledge}, e.g., `the sky is blue', `Paris is in France'; \textit{taxonomic knowledge}, e.g., `football players are athletes', `cats are mammals'; \textit{relational knowledge}, e.g., `the nose is part of the skull', `handwriting requires a hand and a writing instrument'; \textit{procedural commonsense}, which includes \text{prescriptive knowledge}, e.g., `one needs an oven before baking cakes', `the electricity should be off while the switch is being repaired' \cite{hobbs1987commonsense}; \textit{sentiment knowledge}, e.g., `rushing to the hospital makes people worried', `being in vacation makes people relaxed'; and \textit{metaphorical knowledge} (e.g., `time flies', `raining cats and dogs'). We believe that it is important to identifiy the most appropriate commonsense knowledge type required for specific tasks, in order to get better downstream performance. Once the knowledge type is identified, we can then select the appropriate knowledge-base(s), and the suitable neural integration mechanisms (e.g., attention-based injection, pre-training, or auxiliary training objectives). 

Accordingly, in this work we conduct a comparison study of different knowledge bases and knowledge integration methods, and we evaluate model performance on two multiple-choice QA datasets that explicitly require commonsense reasoning. In particular, we used ConceptNet \cite{speer2016conceptnet} and the recently-introduced ATOMIC \cite{sap2019atomic} knowledge resources, integrating them with the \textit{Option Comparison Network} model (OCN; \citet{DBLP:journals/corr/abs-1903-03033}), a recent state-of-the-art model for multiple choice QA tasks. We evalutate our models on the \texttt{DREAM} \cite{TACL1534} and \texttt{CommonsenseQA} \cite{talmor-etal-2019-commonsenseqa} datasets. An example from \texttt{DREAM} that requires commonsense is shown in Table \ref{dream-example}, and an example from \texttt{CommonsenseQA} is shown in Table \ref{csqa-example}. Our experimental results and analysis suggest that attention-based injection is preferable for knowledge integration and that the degree of domain overlap, between knowledge-base and dataset, is vital to model success.\footnote{From a terminological standpoint, `domain overlap' here must be interpreted as the overlap between question types in the targeted datasets, and types of commonsense represented in the knowledge bases under consideration.} 

\begin{table}[h]
\footnotesize
\begin{center}
\begin{tabular}{|l|}
\hline \bf Dialogue: \\ 
\textbf{M:} I hear you drive a long way to work every day. \\
\textbf{W:} Oh, yes. it's about sixty miles. but it doesn't seem \\ 
that far, the road is not bad, and there's not much traffic.\\
\textbf{Question:} \\
How does the woman feel about driving to work? \\
\textbf{Answer choices:}\\ 
A. She doesn't mind it as the road conditions are good.\textbf{*} \\
B. She is unhappy to drive such a long way everyday.\\
C. She is tired of driving in heavy traffic. \\
\hline
\end{tabular}
\end{center}
\caption{An example from the \texttt{DREAM} dataset; the asterisk (\textbf{*}) denotes the correct answer.}
\label{dream-example}
\end{table}

\begin{table}[h]
\footnotesize
\begin{center}
\begin{tabular}{|l|}
\hline \bf Question: \\ 
A revolving door is convenient for two direction travel, \\
but it also serves as a security measure at a what? \\
\textbf{Answer choices:}\\ 
A. Bank\textbf{*}; B. Library; C. Department Store; \\D. Mall; E. New York \\
\hline
\end{tabular}
\end{center}
\caption{An example from the \texttt{CommonsenseQA} dataset; the asterisk (\textbf{*}) denotes the correct answer.}
\label{csqa-example}
\end{table}

\section{Related Work}
It has been recognized that many recent QA tasks require external knowledge or commonsense to solve, and numerous efforts have been made in injecting commonsense in neural models. 
\citet{bauer-etal-2018-commonsense} introduced a pipeline for extracting grounded multi-hop commonsense relation paths from ConceptNet and proposed to inject commonsense knowledge into neural models' intermediate representations, using attention.
Similarly, \citet{mihaylov-frank-2018-knowledgeable} also proposed to extract relevant knowledge triples from ConceptNet and use Key-Value Retrieval \cite{miller-etal-2016-key} to gather information from knowledge to enhance the neural representation. 
\citet{DBLP:journals/corr/abs-1809-03568} proposed to pre-train a scoring function using knowledge triples from ConceptNet, to model the direct and indirect relation between concepts. This scoring function was then fused with QA models to make the final prediction.
\citet{DBLP:journals/corr/abs-1902-00993} introduced an entity discovery and linking system to identify the most salient entities in the question and answer-options. Wikipedia abstracts of these entities are then extracted and appended to the reference documents to provide additional information.
\citet{Weissenborn2018DynamicIO} proposed a strategy of dynamically refining word embeddings by reading input text as well as external knowledge, such as ConceptNet and Wikipedia abstracts.  
More recently, \citet{Lin2019KagNetKG} proposed to extract sub-graphs from ConceptNet and embed the knowledge using Graph Convolutional Networks \cite{DBLP:journals/corr/KipfW16}. Then the knowledge representation is integrated with word representation through an LSTM layer and hierarchical attention mechnism. 
\citet{Lv2019GraphBasedRO} introduced graph-based reasoning modules that takes both ConceptNet knowledge triples and Wikipedia text as inputs to refine word representations from a pretrained language model and make predictions. 

Commonsense knowledge integration has also received a lot of attention on many other tasks. \citet{tandon-etal-2018-reasoning} proposed to use commonsense knowledge as hard/soft constraints to bias the neural model's prediction on a procedural text comprehension task. \citet{Ma2018TargetedAS} proposed to used embedded affective commonsense knowledge inside LSTM cell to control the information flow in each gate for sentiment analysis task.  \citet{li-srikumar-2019-augmenting} presented a framework to convert declarative knowlegde into first-order logic that enhance neural networks' training and prediction. \citet{Peters2019KnowledgeEC} and \citet{Levine2019SenseBERTDS} both tried to injecting knowlegde into language models by pretraining on knowledge bases. 

Previous works only focus on using external knowledge sources to improve model performance on certain tasks, disregarding the type of commonsense knowledge and how the domain of the knowledge resource affects results on downstream tasks. In this paper, we examine the roles of knowledge-base domain and specific integration mechanisms on model performance. 

\section{Approach Overview}
\begin{figure*}
    \centering
    \includegraphics[scale=0.4]{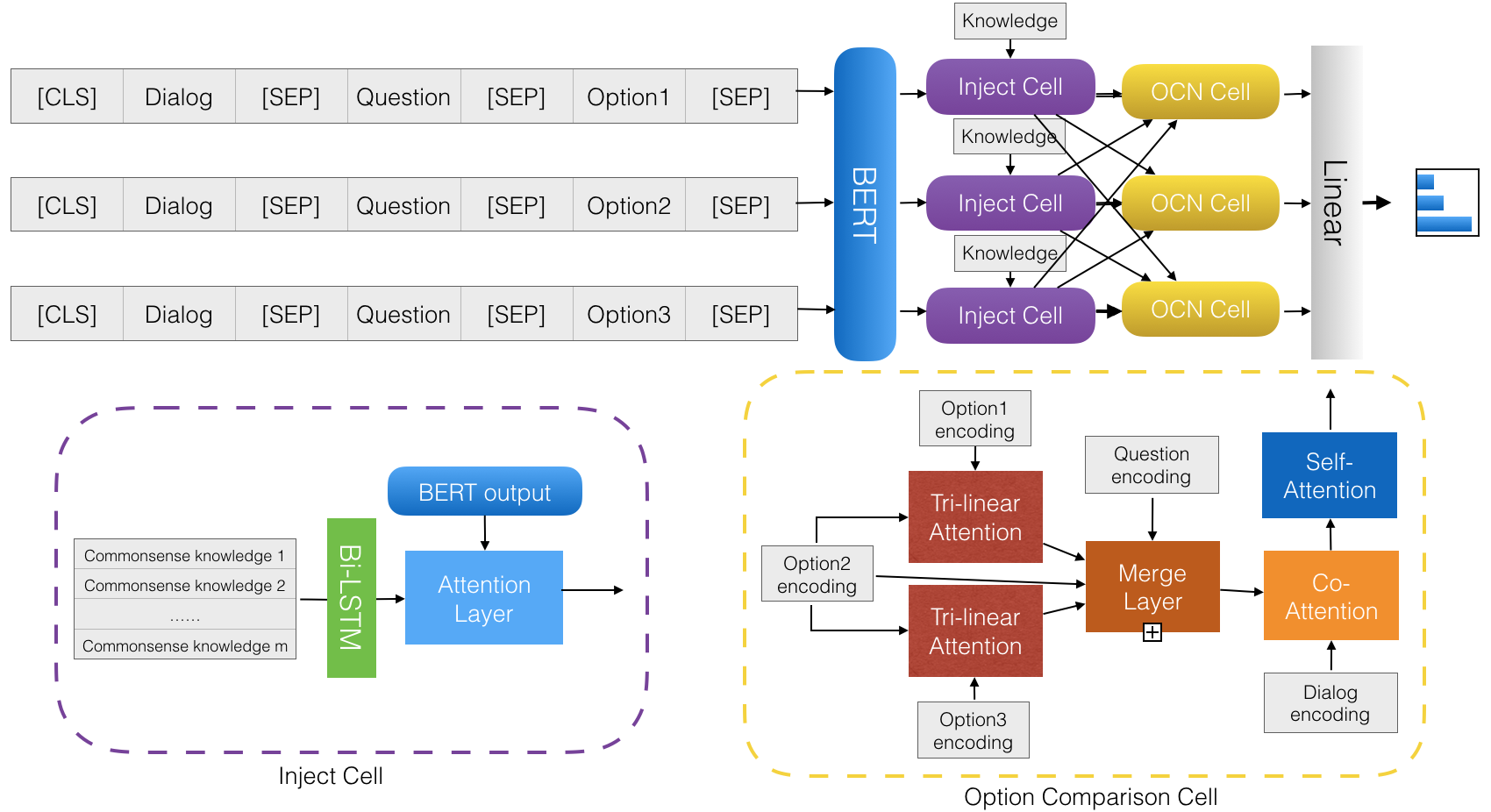}
    \caption{Option Comparison Network with Knowledge Injection}
    \label{fig:model}
\end{figure*}

In this section, we describe the model architecture used in our experiments. Next, we introduce two popular knowledge resources, we define our knowledge-extraction method, then we illustrate various neural knowledge-integration mechanisms.

\subsection{Model architecture}

The BERT model \cite{devlin-etal-2019-bert} has been applied to numerous QA tasks and has achieved very promising performance, particularly on the \texttt{DREAM} and \texttt{CommonsenseQA} datasets. 
When utilizing BERT on multiple-choice QA tasks, the standard approach is to concatenate the dialogue context and the question with each answer-option, in order to generate a list of tokens which is then fed into BERT encoder; a linear layer is added on top, in order to predict the answer. One aspect of this strategy is that each answer-option is encoded independently: from a cognitive perspective, this aspect contradicts how humans typically solve multiple-choice QA tasks, namely by \textit{weighing} each option to find correlations within them, in addition to correlations with respect to the question. To address this issue, \citet{DBLP:journals/corr/abs-1903-03033} introduced the \textit{Option Comparison Network} (OCN) that explicitly models pairwise answer-option interactions, making OCN better-suited for multiple-choice QA task structures. We re-implemented OCN while keeping BERT as its upstream encoder.\footnote{Because the newly-released XLNet has out-performed BERT on various tasks, we considered using XLNet as the OCN's encoder. However, from our initial experiments, XLNet is very unstable, in that it easily provides degenerate solutions---a problem noted by \citet{devlin-etal-2019-bert} for small datasets. We found BERT to be more stable in our study.} Specifically, given a dialogue $D$, a question $Q$, and an answer-option $O_k$, we concatenate them and encode with BERT to get hidden representation $T_{enc} \in \mathbb{R}^{n \times d}$: 
\begin{align}
\label{eq_bert}
    T_{enc} = &~\text{BERT}(D;Q;O_k) 
\end{align}
Where $d$ is the size of BERT's hidden representation and $n$ is the total number of words. Next, the dialogue encoding $D_{enc} \in \mathbb{R}^{n_d \times d}$, question encoding $Q_{enc} \in \mathbb{R}^{n_q \times d}$, and answer-option encoding $O_{k,enc} \in \mathbb{R}^{n_o \times d}$
are separated from $T_{enc}$. Here, option-encoding consists both of question and option, i.e. $Q_{enc}$ $\subseteq$ $O_{k,enc}$ and $n_d + n_o = n$, as suggested by \citet{DBLP:journals/corr/abs-1903-03033}. Given a set of options $O_k$ ($k=1,2,...$), these options are compared, pairwise, using standard tri-linear attention \cite{Seo2016BidirectionalAF}:
\begin{align}
\label{tri_att}
    \text{Att}(u, v) =  W_1 \cdot u + W_2 \cdot & v + (W_3 \circ  v) \cdot u 
\end{align}
Where, $W_1, W_2, W_3 \in \mathbb{R}^{d}$ are trainable weights and $u \in \mathbb{R}^{x \times d}$, $v \in \mathbb{R}^{y \times d}$ are input matrices; $x$ and $y$ here are generic placeholder for input lengths; matrix multiplication and elementwise multiplication are denoted by ($\cdot$) and ($\circ$), respectively. Next, we gather information from all other options, to form a new option representation $O_{k,new} \in \mathbb{R}^{n_o \times d}$. Formally, given option $O_{k,enc}$ and another option $O_{l,enc} \in \mathbb{R}^{n_l \times d}$, $O_{k,new}$ is computed as follows:
\begin{align}
    O_k^l = &~O_{l,enc} \cdot \text{Att}(O_{l,enc}, O_{k,enc})\\
    \widetilde{O_k^l} = &~[O_{k,enc}-O_k^l; O_{k,enc} \circ O_k^l]\\
    O_{k,new} = &~\text{tanh}(W_c \cdot [O_{k,enc};\{\widetilde{O_k^l}\}_{l\neq k} ])
\end{align}
Where, $W_c \in \mathbb{R}^{(d+2d(|O|-1)) \times d}$, $|O|$ denotes total number of options and $n_l$ denotes the number of words in the compared option. Then, a gating mechanism is used to fuse the option-wise correlation information $O_{k,new}$ with the current option-encoding $O_{k,enc}$. Gating values are computed as:  
\begin{align}
    G = &~\text{sigmoid}(W_g[O_{k,enc}; O_{k,new}; \widetilde{Q}]) \\ 
    \widetilde{Q} = &~Q_{enc} \cdot \text{softmax}(Q_{enc} \cdot V_a)^T \\
    O_{fuse} = &~G \circ O_{k,enc} + (1 - G) \circ O_{k,new}
\end{align}
Here, $W_g \in \mathbb{R}^{3d \times d}$ and $V_a \in \mathbb{R}^{d \times 1}$. Co-attention \cite{Xiong2016DynamicCN} is applied to re-read the dialogue, given the fused option-correlation features:
\begin{align}
    A_{do} = &~\text{Att}(D_{enc}, O_{fuse}) \\
    A_{od} = &~\text{Att}(O_{fuse}, D_{enc}) \\ 
    O_d = A_{od} & \cdot [D_{enc}; A_{do} \cdot O_{fuse}] \\
    \widetilde{O_{d}} = &~\text{ReLU}(W_p([O_d; O_{fuse}]))
\end{align}
Here, $W_p \in \mathbb{R}^{3d \times d}$. Finally, self-attention \cite{wang-etal-2017-gated} is used to compute final option representation $\widetilde{O_f} \in \mathbb{R}^{n_o \times d}$:  
\begin{align}
     O_s = &~\widetilde{O_{d}} \cdot \text{Att}(\widetilde{O_{d}}, \widetilde{O_{d}}) \\ 
    O_f = [ \widetilde{O_{d}}; O_s, & \widetilde{O_{d}} - O_s;  \widetilde{O_{d}} \circ O_s] \\
    \widetilde{O_f} = &~\text{ReLU}(W_f \cdot O_f)
\end{align}

Unlike the vanilla BERT model, which takes the first token to predict the answer, max-pooling is applied on the sequence dimension of $\widetilde{O_f} \in \mathbb{R}^{n_o \times d}$, in order to generate the final prediction.

\subsection{Knowledge bases}
The first knowledge-base we consider for our experiments is ConceptNet \cite{speer2016conceptnet}. ConceptNet contains over 21 million edges and 8 million nodes (1.5 million nodes in the partition for the English vocabulary), generating triples of the form $(C1, r, C2)$: the natural-language concepts $C1$ and $C2$ are associated by commonsense relation $r$, e.g., \textit{(dinner, AtLocation, restaurant)}. Thanks to its coverage, ConceptNet is one of the most popular semantic networks for commonsense. ATOMIC \cite{sap2019atomic} is a new knowledge-base that focuses on procedural knowledge. Triples are of the form \textit{(Event, r, \{Effect$|$Persona$|$Mental-state\}}), where head and tail are short sentences or verb phrases and $r$ represents an \textit{if-then} relation type. An example would be: \textit{(X compliments Y, xIntent, X wants to be nice)}. Since both \texttt{DREAM} and \texttt{CommonsenseQA} datasets are open-domain and require general commonsense, we think these knowledge-bases are most appropriate for our investigation. 

\subsection{Knowledge elicitation}
\textbf{ConceptNet}. For the \texttt{DREAM} dataset, we find ConceptNet relations that connect dialogues and questions to the answer-options. The intuition is that these relation paths would provide explicit evidence that would help the model find the answer. Formally, given a dialogue $D$, a question $Q$, and an answer-option $O$, we find all ConceptNet relations \textit{(C1, r, C2)}, such that $C1 \in (D+Q)$ and $C2 \in O$, or vice versa. This rule works well for single-word concepts. However, a large number of concepts in ConceptNet are actually phrases, and finding exactly matching phrases in $D/Q/O$ is much harder. To fully utilize phrase-based ConceptNet relations, we relaxed the exact-match constraint to the following:
\begin{equation}
\frac{\text{\# words in C $\cap$ S}}{\text{\# words in C}} > 0.5 
\end{equation}
Here, $S$ represents $D/Q/O$, depending on which sequence we try to match the concept $C$ to. Additionally, when the part-of-speech (POS) tag for a concept is available, we make sure it matches the POS tag of the corresponding word in $D/Q/O$. For \texttt{CommonsenseQA}, we use the same procedure to find ConceptNet relations for each answer-option, except that only $Q$ is present and used. Table \ref{csqa-cn} shows the extracted ConceptNet triples for the \texttt{CommonsenseQA} example in Table \ref{csqa-example}. It is worth noting that we are able to extract the original ConceptNet sub-graph that was used to create the question, along with some extra triples. Although not perfect, the bold ConceptNet triple does provide some clue that could help the model resolve the correct answer.

\begin{table*}[h]
\footnotesize
\begin{center}
\begin{tabular}{|c|l|}
\hline 
\bf Options & \bf Extracted ConceptNet triples \\
\hline
Bank & (revolving door \textit{AtLocation} bank) \bf (bank RelatedTo security) \\
Library & (revolving door \textit{AtLocation} library) \\
Department Store & (revolving door \textit{AtLocation} store) (security IsA department) \\
Mall & (revolving door \textit{AtLocation} mall) \\
New York & (revolving door \textit{AtLocation} New York) \\
\hline
\end{tabular}
\end{center}
\caption{Extracted ConceptNet relations for sample shown in Table \ref{csqa-example}.}
\label{csqa-cn}
\end{table*}

\begin{table*}[h]
\footnotesize
\begin{center}
\begin{tabular}{|c|l|}
\hline 
\bf Input sentence & \bf Generated ATOMIC relations \\
\hline
Utterance 1 & (xAttr dedicated) (xWant to get to work) \\
Utterance 2 & (xAttr far) \textbf{(xReact happy)} (xWant to get to their destination) \\
Option A & \textbf{(xAttr calm)} (xWant to avoid the road)  \\
Option B & \textbf{(xAttr careless)} (xReact annoyed) (xEffect get tired) \\
Option C & \textbf{(xAttr frustrated)} (xEffect get tired) (xWant to get out of car) \\
\hline
\end{tabular}
\end{center}
\caption{Sample generated ATOMIC relations for sample shown in Table \ref{dream-example}.}
\label{dream-at}
\end{table*}

\noindent
\textbf{ATOMIC}. We observe that many questions in \texttt{DREAM} inquire about agent's opinion and feeling. Superficially, this particular question type seems well-suited for ATOMIC, whose focus is on folk psychology and related general implications; we could frame our goal as evaluating whether ATOMIC can provide relevant knowledge to help answer these questions. However, one challenge to this strategy is that heads and tails of knowledge triples in ATOMIC are short sentences or verb phrases, while rare words and person-references are reduced to blanks and PersonX/PersonY, respectively. This calls for a new matching procedure, different from the ConceptNet extraction strategy, for eliciting ATOMIC-specific relations: we rely on the recently-published COMET model \cite{bosselut-etal-2019-comet} to generate new ATOMIC relations, with intermediate phrasal resolutions. In particular, we first segmented all dialogues, questions, and answer-options into sentences. We further segment long sentences into sub-sentences, using commas as seperators. Because only verb-phrases satisfy the definition of an ``event" in ATOMIC (i.e., relations are only invoked by verbs), we remove all sentences/sub-sentences that do not contain any verb. Next, we use a pre-trained COMET model \cite{bosselut-etal-2019-comet} to generate all possible ATOMIC relations, for all candidate sentences/sub-sentences and we use greedy-decoding to take the 1-best sequences. Table \ref{dream-at} shows the sample ATOMIC relations, generated using the \texttt{DREAM} example in Table \ref{dream-example}. It is interesting to note that the reaction for the woman agent (second utterance) is identified as \textit{happy}, since she said that `the road is not bad.' If we compare the identified attributes for answer-options, the one from correct answer seems to be sematically closer than the other two.

\subsection{Knowledge injection}
Given previously extracted/generated knowledge triples, we need to integrate them with the OCN model. Inspired by \citet{bauer-etal-2018-commonsense}, we propose to use attention-based injection. For ConceptNet knowledge triples, we first convert concept-relation tokens into regular tokens, in order to generate a pseudo-sentence. For example, ``\textit{(book, AtLocation, library)}" would be converted to ``book at location library." Next, we use the BERT embedding layer to generate an embedding of this pseudo-sentence, with $C$ denoting a ConceptNet relation:
\begin{equation}
    H_{C} = \text{BiLSTM}(C) 
\end{equation}
If we let $H_{C} \in \mathbb{R}^{1\times2l} $ be the concatenation of the final hidden states and $l$ be the number of hidden units in the LSTM layer, then $m$ ConceptNet relations would yield the commonsense knowledge matrix $H_{M} \in \mathbb{R}^{m\times2l} $. We adopt the attention mechanism used in QAnet \cite{46691} to model the interaction between $H_{M}$ and the BERT encoding output $T_{enc}$ (from Equation \ref{eq_bert}): 
\begin{align}
     \widetilde{H}_{M} = &~H_{M} \cdot W_{proj} \\
     \mathcal{S} = &~\text{Att}({H}_{M}, T_{enc})\\
     A_m = &~\text{softmax}(\mathcal{S}) \cdot \widetilde{H}_{M} \\ 
     A_t = \text{softmax}(\mathcal{S}) \cdot & \text{softmax}(\mathcal{S}^T) \cdot T_{enc} \\
     T_C = [T_{enc}; A_m; & T_{enc} \circ A_m; T_{enc} \circ A_t]\\ 
     T_{out} = &~\text{ReLU}(T_C \cdot W_a)
\end{align}
Specifically, $H_{M}$ is first projected into the same dimension as $T_{enc}$, using $W_{proj} \in \mathbb{R}^{2l \times d}$. Then, the similarty matrix $\mathcal{S} \in \mathbb{R}^{n \times m}$ is computed using tri-linear attention, as in Equation \ref{tri_att}. We then use $\mathcal{S}$ to compute text-to-knowledge attention $A_m \in \mathbb{R}^{n \times d}$ and knowledge-to-text attention $A_t \in \mathbb{R}^{n \times d}$. Finally, the knowledge-aware textual representation $T_{out} \in \mathbb{R}^{n \times d}$ is computed, where $W_a \in \mathbb{R}^{4d \times d}$. $T_{out}$ is fed to subsequent layers (in place of $T_{enc}$), in order to generate the prediction. The model structure with knowledge-injection is summarized in Figure \ref{fig:model}. 

For ATOMIC knowledge triples, the injection method is slightly different. Because heads of these knowledge triples are sentences/utterances and the tails contain attributes of the persons (i.e., subject and object of the sentence), it is not possible to directly inject the knowledge triples, as-is. We replace the heads of the ATOMIC knowledge triples with the corresponding speaker for dialogues and leave as blank for the answer-options. Next, we convert the special relation tokens into regular tokens, e.g., ``xIntent''$\Rightarrow$``intent" and ``oEffect"$\Rightarrow$ ``others effect'', to make pseudo-sentences. As a result, an ATOMIC relation ``\textit{(the road is not bad, xReact, happy)}" would be converted to ``\textit{(W, react, happy)}." Moreover, as the ATOMIC knowledge triples are associated with dialogues and answer-options, independently, we inject option relations into $O_{enc} \in \mathbb{R}^{n_o \times d}$ and dialogue relations into $D_{enc}$, respectively, using the injection method described above.  

\linespread{0.4}
\subsection{Knowledge pre-training}
Pre-training large-capacity models (e.g., BERT, GPT \cite{radford2019language}, XLNet \cite{yang2019xlnet}) on large corpora, then fine-tuning on more domain-specific information, has led to performance improvements on various tasks. Inspired by this, our goal in this section is to observe the effect of pre-training BERT on commonsense knowledge and refining the model on task-specific content from our \texttt{DREAM} and \texttt{CommonsenseQA} corpora. Essentially, we would like to test if pre-training on our external knowledge resources can help the model acquire commonsense. For the ConceptNet pre-training procedure, 
pre-training BERT on pseudo-sentences formulated from ConceptNet knowledge triples does not provide much gain on performance. Instead, we trained BERT on the \textit{Open Mind Common Sense} (OMCS) corpus \cite{Singh:2002:OMC:646748.701499}, the original corpus that was used to create ConceptNet. We extracted about 930K English sentences from OMCS and randomly masked out 15\% of the tokens; we then fine-tuned BERT, using a masked language model objective. Then we load this fine-tuned model into OCN and trained on \texttt{DREAM} and \texttt{CommonsenseQA} tasks.  
As for pre-training on ATOMIC, we again use COMET to convert ATOMIC knowledge triples into sentences; we created special tokens for 9 types of relations as well as blanks. Next, we randomly masked out 15\% of the tokens, only masking out tail-tokens. We use the same OMCS pre-training procedure. 
\begin{table}[h]
\footnotesize
\begin{center}
\begin{tabular}{|c|cc|}
\hline \bf Models & \bf Dev Acc & \bf Test Acc \\ \hline
BERT Large(*) & 66.0 & 66.8 \\ 
XLNet(*) & - & \bf 72.0 \\ \hline
OCN & 70.0 & 69.8 \\
OCN + CN injection & \bf 70.5 & 69.6 \\
OCN + AT injection & 69.6 & \bf 70.1 \\
OCN + OMCS pre-train & 64.0 & 62.6 \\
OCN + ATOMIC pre-train & 60.3 & 58.8 \\
\hline
\end{tabular}
\end{center}
\caption{Results on \texttt{DREAM}; the asterisk (*) denotes results taken from leaderboard.}
\label{dream-results}
\end{table}

\begin{table}[h]
\footnotesize
\begin{center}
\begin{tabular}{|c|c|}
\hline \bf Models & \bf Dev Acc  \\ \hline
BERT + OMCS pre-train(*) & 68.8 \\
RoBERTa + CSPT(*) & \bf 76.2 \\ \hline
OCN & 64.1 \\
OCN + CN injection & 67.3 \\
OCN + OMCS pre-train & 65.2 \\
OCN + ATOMIC pre-train & 61.2 \\
OCN + OMCS pre-train + CN inject & \bf 69.0 \\

\hline
\end{tabular}
\end{center}
\caption{Results on \texttt{CommonsenseQA}; the asterisk (*) denotes results taken from leaderboard.}
\label{csqa-results}
\end{table}

\section{Experiments}
\label{experiments}

\subsection{Datasets}
\label{datasets}
We choose to evaluate our hypotheses using the \texttt{DREAM} and \texttt{CommonsenseQA} datasets, because some / all questions require commonsense reasoning and because there remains a large gap between state-of-the-art models and human performance. 
  
\texttt{DREAM} is a dialogue-based multiple-choice QA dataset, introduced by \citet{TACL1534}. It was collected from English-as-a-foreign-language examinations, designed by human experts. The dataset contains 10,197 questions for 6,444 dialogues in total, and each question is associated with 3 answer-options. The authors point out that 34\% of questions require commonsense knowledge to answer, which includes social implication, speaker's intention, or general world knowledge. 

\texttt{CommonsenseQA} is a multiple-choice QA dataset that specifically measure commonsense reasoning \cite{talmor-etal-2019-commonsenseqa}. This dataset is constructed based on ConceptNet \cite{speer2016conceptnet}. Specifically, a source concept is first extracted from ConceptNet, along with 3 target concepts that are connected to the source concept, i.e., a sub-graph. Crowd-workers are then asked to generate questions, using the source concept, such that only one of the target concepts can correctly answer the question. Additionally, 2 more distractor concepts are selected by crowd-workers so that each question is associated with 5 answer-options. In total, the dataset contains 12,247 questions. For \texttt{CommonsenseQA}, we evaluate models on the development-set only, since test-set answers are not publicly available.

\subsection{Training details}
For ease of comparison, we borrow hyperparameter settings from \citet{pan2019improving}; we used the BERT Whole-Word Masking Uncased model \cite{devlin2018bert} for all experiments. For \texttt{DREAM} experiments, we used a max sequence-length of 512, batch-size of 24, learning rate of $1e^{-5}$, and we trained the model for 16 epochs. For \texttt{CommonsenseQA}, we used a max sequence length of 60, batch-size of 32, learning rate of $1e^{-5}$, and trained for 8 epochs. For pre-training on OMCS, we used max sequence length of 35, batch-size of 32, learning rate of $3e^{-5}$, and trained for 3 epochs. For pre-training on ATOMIC, the max sequence length is changed to 45, other hyperparameters remain the same, and we only use the ATOMIC training set. When using OCN on \texttt{CommonsenseQA}, since there is no dialogue, we compute co-attention with $Q_{enc}$, in place of $D_{enc}$, in order to keep the model structure consistent.

\subsection{Results}
\texttt{DREAM} results are shown in Table \ref{dream-results}, and \texttt{CommonsenseQA} results are shown in Table \ref{csqa-results}. For all of our experiments, we run 3 trials with different random seeds and we report average scores in the tables. Evaluated on \texttt{DREAM}, our OCN model got a significant performance boost (+3.0\%), compared to BERT-large from previous work. We think the reasons are that OCN is better-suited for the task and that we used BERT Whole-Word Masking Uncased model. OCN with ConceptNet knowledge-injection achieves slightly better results on the development-set, while ATOMIC knowledge-injection helps achieve a small improvement on the test-set. However, we recognize that these improvements are very limited; to our surprise, OCN pre-trained on OMCS or ATOMIC got significantly lower performance.

As for results on \texttt{CommonsenseQA}, ConceptNet knowledge-injection provides a significant performance boost (+2.8\%), compared to the OCN baseline, suggesting that explicit links from question to answer-options help the model find the correct answer.
Pre-training on OMCS also provides a small performance boost to the OCN baseline. Since both ConceptNet knowledge-injection and OMCS pre-training are helpful, we combine both approaches with OCN and we are able to achieve further improvement (+4.9\%).  
Finally, similar to the results on \texttt{DREAM}, OCN pre-trained on ATOMIC yields a siginificant performance drop.

\begin{table*}[h]
\footnotesize
\begin{center}
\adjustbox{max width=\textwidth}{
\begin{tabular}{|c|ccccccc|}
\hline \bf Models & \bf M(54) & \bf S(15) & \bf A+L(11) & \bf L(228) & \bf C+L(122) & \bf C(14) & \bf C+S(60) \\ \hline
OCN & 88.9 & 86.7 & 27.3 & 75.9 & 60.7 & 71.4 & 70.0\\
OCN + CN injection & 83.3(-5.6) & 86.7(+0.0) & 18.2(-9.2) & 76.8(+0.9) & 59.8(-0.9)  & 64.3(-7.1)  & 78.3(\textbf{+8.3}) \\
OCN + AT injection & 88.9(+0.0) & 80.0(-6.7) & 27.3(+0.0) & 75.9(+0.0) & 66.4(\textbf{+5.7})  & 71.4(+0.0) & 75(\textbf{+5.0})\\
OCN + OMCS pre-train & 70.4(-18.5) & 73.3(-13.4) & 45.4(+18.1) & 69.7(-6.2) & 48.4(-12.3)  & 57.1(-14.3) & 68.3(\textbf{-1.7})\\
OCN + ATOMIC pre-train & 66.6(-22.3) & 86.7(+0.0) & 18.2(-9.2) & 64.0(-11.9) & 51.6(-9.1)  & 42.9(-28.5) & 70.0(\textbf{+0.0}) \\
\hline
\end{tabular}}
\end{center}
\caption{Accuracies for each \texttt{DREAM} question type: \textbf{M} means \textit{Matching}, \textbf{S} means \textit{Summary}, \textbf{L} means \textit{Logic inference}, \textbf{C} means \textit{Commonsense inference}, and \textbf{A} means \textit{Arithmatic inference}. Numbers beside types denote the number of questions of that type.}
\label{dream-errors1}
\end{table*}

\begin{table*}[h]
\footnotesize
\begin{center}
\adjustbox{max width=\textwidth}{
\begin{tabular}{|c|cccccccc|}
\hline \bf Models & \bf AtLoc.(596) & \bf Cau.(194) & \bf Cap.(109) & \bf Ant.(92) & \bf H.Pre.(46) & \bf H.Sub.(39) & \bf C.Des.(28) & \bf Des.(27) \\ \hline
OCN & 64.9 & 66.5 & 65.1 & 55.4 & 69.6 & 64.1 & 57.1 & 66.7\\
+CN inj, & 67.4(+2.5) & 70.6(+4.1) & 66.1(+1.0) & 60.9(+5.5) & 73.9(+4.3)  & 66.7(+2.6) & 64.3(+7.2) & 77.8(+11.1)\\
+OMCS & 68.8(+3.9) & 63.9(-2.6) & 62.4(-2.7) & 60.9(+5.5) & 71.7(+2.1)  & 59.0(-5.1) & 64.3(+7.2) & 74.1(+7.4)\\
+ATOMIC & 62.8(-2.1) & 66.0(\textbf{-0.5}) & 60.6(-4.5) & 52.2(-3.2) & 63.0(-6.6)  & 56.4(-7.7) & 60.7(\textbf{+3.6}) & 74.1(\textbf{+7.4}) \\
+OMCS+CN & 71.6(+6.7) & 71.6(+5.1) & 64.2(+0.9) & 59.8(+4.4) & 69.6(+0.0)  & 69.2(+5.1) & 75.0(+17.9) & 70.4(+3.7) \\
\hline
\end{tabular}}
\end{center}
\caption{Accuracies for each \texttt{CommonsenseQA} question type: \textbf{AtLoc.} means \textit{AtLocation}, \textbf{Cau.} means Causes, \textbf{Cap.} means \textit{CapableOf}, \textbf{Ant.} means \textit{Antonym}, \textbf{H.Pre.} means \textit{HasPrerequiste}, \textbf{H.Sub} means \textit{HasSubevent}, \textbf{C.Des.} means \textit{CausesDesire}, and \textbf{Des.} means \textit{Desires}. Numbers beside types denote the number of questions of that type.}
\label{csqa-errors1}
\end{table*}

\section{Error Analysis}
To better understand when a model performs better or worse with knowledge-integration, we analyzed model predictions. \texttt{DREAM} dataset provides annotations for about 1000 questions: ~500 questions in the development-set and ~500 in the test-set. Specifically, questions are manually classified into 5 categories: Matching, Summary, Logic inference, Commonsense inference, and Arithmetic inference; and each question can be classified under multiple categories. We refer readers to \citet{TACL1534} for additional category information. We extracted model predictions for these annotated questions in test-set and grouped them by types. The accuracies for each question-group are shown in Table \ref{dream-errors1}. Note that we omitted 2 categories that have less than 10 questions. For the ConceptNet and the ATOMIC knowledge-injection models, we can see that they did better on questions that involve commonsense (last 3 columns in the table), and the performance on other types are about the same or slightly worse, compared to baseline OCN. As for models pre-trained on OMCS corpus or ATOMIC knowledge-base, we already saw that these model performances drop, compared to the baseline. When we look at the performance difference in each question type, it is clear that some categories account for the performance drop more than others. For example, for both the OMCS pre-trained model and the ATOMIC pre-trained model, performance drops significantly for Matching questions, in particular. On the other hand, for questions that require both commonsense inference and summarization, both models' performances only dropped slightly or did not change. Based on these results, we infer that commonsense knowledge-injection with attention is making an impact on models' weight distributions. The model is able to do better on questions that require commonsense but is losing performance on other types, suggesting a direction for future research in developing more robust (e.g., conditional) injection methods. Moreover, pre-training on knowledge-bases seems to have a larger impact on models' weight distributions, resulting in inferior performance. This weight distribution shift also favors of commonsense, as we see that commonsense types are not affected as much as other types. 
We also conducted similar analysis for \texttt{CommonsenseQA}. Since all questions in \texttt{CommonsenseQA} require commonsense reasoning, we classify questions based on the ConceptNet relation between the question concept and correct answer concept. The intuition is that the model needs to capture this relation in order to answer the question. The accuracies for each question type are shown in Table \ref{csqa-errors1}. Note that we have omitted question types that have less than 25 questions. We can see that with ConceptNet relation-injection, all question types got performance boosts, for both OCN model and OCN pre-trained on OMCS, suggesting that knowledge is indeed helpful for the task. In the case of OCN pre-trained on ATOMIC, although the overall performance is much lower than OCN baseline, it is interesting to see that performance for the ``Causes" type is not significantly affected. Moreover, performance for ``CausesDesire" and ``Desires" types actually got much better. As noted by \cite{sap2019atomic}, ``Causes" in ConceptNet is similar to ``Effects" and ``Reactions" in ATOMIC; and ``CausesDesire" in ConceptNet is similar to ``Wants" in ATOMIC. This result also correlates with our findings from our analysis on \texttt{DREAM}, wherein we found that models with knowledge pre-training perform better on questions that fit knowledge domain but perform worse on others. In this case, pre-training on ATOMIC helps the model do better on questions that are similar to ATOMIC relations, even though overall performance is inferior. Finally, we noticed that questions of type ``Antonym" appear to be the hardest ones. Many questions that fall into this category contain negations, and we hypothesize that the models still lack the ability to reason over negation sentences, suggesting another direction for future improvement. 

\section{Discussion}
Based on our experimental results and error analysis, we see that external knowledge is only helpful when there is alignment between questions and knowledge-base types. Thus, it is crucial to identify the question type and apply the best-suited knowledge. In terms of knowledge-integration methods, attention-based injection seems to be the better choice for pre-trained language models such as BERT. Even when alignment between knowledge-base and dataset is sub-optimal, the performance would not degrade. On the other hand, pre-training on knowledge-bases would shift the language model's weight distribution toward its own domain, greatly. If the task domain does not fit knowledge-base well, model performance is likely to drop. When the domain of the knowledge-base aligns with that of the dataset perfectly, both knowledge-integration methods bring performance boosts and a combination of them could bring further gain.

\section{Future Work}
We have presented a survey on two popular knowledge bases (ConceptNet and ATOMIC) and recent knowledge-integration methods (attention and pre-training), on commonsense QA tasks. Evaluation on two QA datasets suggests that alignment between knowledge-bases and datasets plays a crucial role in knowledge-integration. We believe it is worth conducting a more comprehensive study of datasets and knowledge-bases and putting more effort towards defining an auxiliary learning objective, in a constrained-optimization (i.e., multi-task learning) framework, that identifies the type of knowledge required, based on data characteristics. In parallel, we are also interested in building a \textit{global commonsense knowledge base} by aggregating ConceptNet, ATOMIC, and potentially other resources like FrameNet \cite{baker1998berkeley} and MetaNet \cite{dodge2015metanet}, on the basis of a shared-reference ontology (following the approaches described in \cite{gangemi2010interfacing} and \cite{scheffczyk2010reasoning}): the goal would be to assess whether injecting knowledge structures from a semantically-cohesive lexical knowledge base of commonsense guarantees stable model accuracy across datasets. 


\bibliography{emnlp-ijcnlp-2019}

\begin{thebibliography}{45}
\expandafter\ifx\csname natexlab\endcsname\relax\def\natexlab#1{#1}\fi

\bibitem[{Baker et~al.(1998)Baker, Fillmore, and Lowe}]{baker1998berkeley}
Collin~F Baker, Charles~J Fillmore, and John~B Lowe. 1998.
\newblock The berkeley framenet project.
\newblock In \emph{Proceedings of the 17th international conference on
  Computational linguistics-Volume 1}, pages 86--90. Association for
  Computational Linguistics.

\bibitem[{Bauer et~al.(2018)Bauer, Wang, and
  Bansal}]{bauer-etal-2018-commonsense}
Lisa Bauer, Yicheng Wang, and Mohit Bansal. 2018.
\newblock \href {https://doi.org/10.18653/v1/D18-1454} {Commonsense for
  generative multi-hop question answering tasks}.
\newblock In \emph{Proceedings of the 2018 Conference on Empirical Methods in
  Natural Language Processing}, pages 4220--4230, Brussels, Belgium.
  Association for Computational Linguistics.

\bibitem[{Bosselut et~al.(2019)Bosselut, Rashkin, Sap, Malaviya, Celikyilmaz,
  and Choi}]{bosselut-etal-2019-comet}
Antoine Bosselut, Hannah Rashkin, Maarten Sap, Chaitanya Malaviya, Asli
  Celikyilmaz, and Yejin Choi. 2019.
\newblock \href {https://www.aclweb.org/anthology/P19-1470} {{COMET}:
  Commonsense transformers for automatic knowledge graph construction}.
\newblock In \emph{Proceedings of the 57th Annual Meeting of the Association
  for Computational Linguistics}, pages 4762--4779, Florence, Italy.
  Association for Computational Linguistics.

\bibitem[{Davis(2014)}]{davis2014representations}
Ernest Davis. 2014.
\newblock \emph{Representations of commonsense knowledge}.
\newblock Morgan Kaufmann.

\bibitem[{Devlin et~al.(2018)Devlin, Chang, Lee, and
  Toutanova}]{devlin2018bert}
Jacob Devlin, Ming-Wei Chang, Kenton Lee, and Kristina Toutanova. 2018.
\newblock Bert: Pre-training of deep bidirectional transformers for language
  understanding.
\newblock \emph{arXiv preprint arXiv:1810.04805}.

\bibitem[{Devlin et~al.(2019)Devlin, Chang, Lee, and
  Toutanova}]{devlin-etal-2019-bert}
Jacob Devlin, Ming-Wei Chang, Kenton Lee, and Kristina Toutanova. 2019.
\newblock \href {https://doi.org/10.18653/v1/N19-1423} {{BERT}: Pre-training of
  deep bidirectional transformers for language understanding}.
\newblock In \emph{Proceedings of the 2019 Conference of the North {A}merican
  Chapter of the Association for Computational Linguistics: Human Language
  Technologies, Volume 1 (Long and Short Papers)}, pages 4171--4186,
  Minneapolis, Minnesota. Association for Computational Linguistics.

\bibitem[{Dodge et~al.(2015)Dodge, Hong, and Stickles}]{dodge2015metanet}
Ellen Dodge, Jisup Hong, and Elise Stickles. 2015.
\newblock Metanet: Deep semantic automatic metaphor analysis.
\newblock In \emph{Proceedings of the Third Workshop on Metaphor in NLP}, pages
  40--49.

\bibitem[{Gangemi et~al.(2010)Gangemi, Guarino, Masolo, and
  Oltramari}]{gangemi2010interfacing}
Aldo Gangemi, Nicola Guarino, Claudio Masolo, and Alessandro Oltramari. 2010.
\newblock Interfacing wordnet with dolce: towards ontowordnet.
\newblock \emph{Ontology and the Lexicon: A Natural Language Processing
  Perspective}, pages 36--52.

\bibitem[{Hermann et~al.(2015)Hermann, Kocisky, Grefenstette, Espeholt, Kay,
  Suleyman, and Blunsom}]{NIPS2015_5945}
Karl~Moritz Hermann, Tomas Kocisky, Edward Grefenstette, Lasse Espeholt, Will
  Kay, Mustafa Suleyman, and Phil Blunsom. 2015.
\newblock \href
  {http://papers.nips.cc/paper/5945-teaching-machines-to-read-and-comprehend.pdf}
  {Teaching machines to read and comprehend}.
\newblock In C.~Cortes, N.~D. Lawrence, D.~D. Lee, M.~Sugiyama, and R.~Garnett,
  editors, \emph{Advances in Neural Information Processing Systems 28}, pages
  1693--1701. Curran Associates, Inc.

\bibitem[{Hobbs et~al.(1987)Hobbs, Croft, Davies, Edwards, and
  Laws}]{hobbs1987commonsense}
Jerry~R Hobbs, William Croft, Todd Davies, Douglas Edwards, and Kenneth Laws.
  1987.
\newblock Commonsense metaphysics and lexical semantics.
\newblock \emph{Computational linguistics}, 13(3-4):241--250.

\bibitem[{Kipf and Welling(2016)}]{DBLP:journals/corr/KipfW16}
Thomas~N. Kipf and Max Welling. 2016.
\newblock \href {http://arxiv.org/abs/1609.02907} {Semi-supervised
  classification with graph convolutional networks}.
\newblock \emph{CoRR}, abs/1609.02907.

\bibitem[{Lai et~al.(2017)Lai, Xie, Liu, Yang, and
  Hovy}]{DBLP:journals/corr/LaiXLYH17}
Guokun Lai, Qizhe Xie, Hanxiao Liu, Yiming Yang, and Eduard~H. Hovy. 2017.
\newblock \href {http://arxiv.org/abs/1704.04683} {{RACE:} large-scale reading
  comprehension dataset from examinations}.
\newblock \emph{CoRR}, abs/1704.04683.

\bibitem[{Levine et~al.(2019)Levine, Lenz, Dagan, Padnos, Sharir,
  Shalev-Shwartz, Shashua, and Shoham}]{Levine2019SenseBERTDS}
Yoav Levine, Barak Lenz, Or~Dagan, Dan Padnos, Or~Sharir, Shai Shalev-Shwartz,
  Amnon Shashua, and Yoav Shoham. 2019.
\newblock Sensebert: Driving some sense into bert.
\newblock \emph{ArXiv}, abs/1908.05646.

\bibitem[{Li and Srikumar(2019)}]{li-srikumar-2019-augmenting}
Tao Li and Vivek Srikumar. 2019.
\newblock \href {https://doi.org/10.18653/v1/P19-1028} {Augmenting neural
  networks with first-order logic}.
\newblock In \emph{Proceedings of the 57th Annual Meeting of the Association
  for Computational Linguistics}, pages 292--302, Florence, Italy. Association
  for Computational Linguistics.

\bibitem[{Lin et~al.(2019)Lin, Chen, Chen, and Ren}]{Lin2019KagNetKG}
Bill~Yuchen Lin, Xinyue Chen, Jamin Chen, and Xiang Ren. 2019.
\newblock Kagnet: Knowledge-aware graph networks for commonsense reasoning.
\newblock \emph{ArXiv}, abs/1909.02151.

\bibitem[{Liu et~al.(2019)Liu, Ott, Goyal, Du, Joshi, Chen, Levy, Lewis,
  Zettlemoyer, and Stoyanov}]{DBLP:journals/corr/abs-1907-11692}
Yinhan Liu, Myle Ott, Naman Goyal, Jingfei Du, Mandar Joshi, Danqi Chen, Omer
  Levy, Mike Lewis, Luke Zettlemoyer, and Veselin Stoyanov. 2019.
\newblock \href {http://arxiv.org/abs/1907.11692} {Roberta: {A} robustly
  optimized {BERT} pretraining approach}.
\newblock \emph{CoRR}, abs/1907.11692.

\bibitem[{Lv et~al.(2019)Lv, Guo, Xu, Tang, Duan, Gong, Shou, Jiang, Cao, and
  Hu}]{Lv2019GraphBasedRO}
Shangwen Lv, Daya Guo, Jingjing Xu, Duyu Tang, Nan Duan, Ming Gong, Linjun
  Shou, Daxin Jiang, Guihong Cao, and Songlin Hu. 2019.
\newblock Graph-based reasoning over heterogeneous external knowledge for
  commonsense question answering.
\newblock \emph{ArXiv}, abs/1909.05311.

\bibitem[{Ma et~al.(2018)Ma, Peng, and Cambria}]{Ma2018TargetedAS}
Yukun Ma, Haiyun Peng, and Erik Cambria. 2018.
\newblock Targeted aspect-based sentiment analysis via embedding commonsense
  knowledge into an attentive lstm.
\newblock In \emph{AAAI}.

\bibitem[{Mihaylov and Frank(2018)}]{mihaylov-frank-2018-knowledgeable}
Todor Mihaylov and Anette Frank. 2018.
\newblock \href {https://doi.org/10.18653/v1/P18-1076} {Knowledgeable reader:
  Enhancing cloze-style reading comprehension with external commonsense
  knowledge}.
\newblock In \emph{Proceedings of the 56th Annual Meeting of the Association
  for Computational Linguistics (Volume 1: Long Papers)}, pages 821--832,
  Melbourne, Australia. Association for Computational Linguistics.

\bibitem[{Miller et~al.(2016)Miller, Fisch, Dodge, Karimi, Bordes, and
  Weston}]{miller-etal-2016-key}
Alexander Miller, Adam Fisch, Jesse Dodge, Amir-Hossein Karimi, Antoine Bordes,
  and Jason Weston. 2016.
\newblock \href {https://doi.org/10.18653/v1/D16-1147} {Key-value memory
  networks for directly reading documents}.
\newblock In \emph{Proceedings of the 2016 Conference on Empirical Methods in
  Natural Language Processing}, pages 1400--1409, Austin, Texas. Association
  for Computational Linguistics.

\bibitem[{Ostermann et~al.(2018)Ostermann, Roth, Modi, Thater, and
  Pinkal}]{ostermann-etal-2018-semeval}
Simon Ostermann, Michael Roth, Ashutosh Modi, Stefan Thater, and Manfred
  Pinkal. 2018.
\newblock \href {https://doi.org/10.18653/v1/S18-1119} {{S}em{E}val-2018 task
  11: Machine comprehension using commonsense knowledge}.
\newblock In \emph{Proceedings of The 12th International Workshop on Semantic
  Evaluation}, pages 747--757, New Orleans, Louisiana. Association for
  Computational Linguistics.

\bibitem[{Pan et~al.(2019{\natexlab{a}})Pan, Sun, Yu, Ji, and
  Yu}]{DBLP:journals/corr/abs-1902-00993}
Xiaoman Pan, Kai Sun, Dian Yu, Heng Ji, and Dong Yu. 2019{\natexlab{a}}.
\newblock \href {http://arxiv.org/abs/1902.00993} {Improving question answering
  with external knowledge}.
\newblock \emph{CoRR}, abs/1902.00993.

\bibitem[{Pan et~al.(2019{\natexlab{b}})Pan, Sun, Yu, Ji, and
  Yu}]{pan2019improving}
Xiaoman Pan, Kai Sun, Dian Yu, Heng Ji, and Dong Yu. 2019{\natexlab{b}}.
\newblock \href {https://arxiv.org/abs/1902.00993v1} {Improving question
  answering with external knowledge}.
\newblock \emph{CoRR}, cs.CL/1902.00993v1.

\bibitem[{Peters et~al.(2019)Peters, Neumann, RobertL.Logan, Schwartz, Joshi,
  Singh, and Smith}]{Peters2019KnowledgeEC}
Matthew~E. Peters, Mark Neumann, IV~RobertL.Logan, Roy Schwartz, Vidur Joshi,
  Sameer Singh, and Noah~A. Smith. 2019.
\newblock Knowledge enhanced contextual word representations.
\newblock \emph{ArXiv}, abs/1909.04164.

\bibitem[{Radford et~al.(2019)Radford, Wu, Child, Luan, Amodei, and
  Sutskever}]{radford2019language}
Alec Radford, Jeff Wu, Rewon Child, David Luan, Dario Amodei, and Ilya
  Sutskever. 2019.
\newblock Language models are unsupervised multitask learners.

\bibitem[{Rajpurkar et~al.(2016)Rajpurkar, Zhang, Lopyrev, and
  Liang}]{rajpurkar-etal-2016-squad}
Pranav Rajpurkar, Jian Zhang, Konstantin Lopyrev, and Percy Liang. 2016.
\newblock \href {https://doi.org/10.18653/v1/D16-1264} {{SQ}u{AD}: 100,000+
  questions for machine comprehension of text}.
\newblock In \emph{Proceedings of the 2016 Conference on Empirical Methods in
  Natural Language Processing}, pages 2383--2392, Austin, Texas. Association
  for Computational Linguistics.

\bibitem[{Ran et~al.(2019)Ran, Li, Hu, and
  Zhou}]{DBLP:journals/corr/abs-1903-03033}
Qiu Ran, Peng Li, Weiwei Hu, and Jie Zhou. 2019.
\newblock \href {http://arxiv.org/abs/1903.03033} {Option comparison network
  for multiple-choice reading comprehension}.
\newblock \emph{CoRR}, abs/1903.03033.

\bibitem[{Sap et~al.(2019)Sap, LeBras, Allaway, Bhagavatula, Lourie, Rashkin,
  Roof, Smith, and Choi}]{sap2019atomic}
Maarten Sap, Ronan LeBras, Emily Allaway, Chandra Bhagavatula, Nicholas Lourie,
  Hannah Rashkin, Brendan Roof, Noah~A Smith, and Yejin Choi. 2019.
\newblock Atomic: An atlas of machine commonsense for if-then reasoning.
\newblock In \emph{AAAI}.

\bibitem[{Scheffczyk et~al.(2010)Scheffczyk, Baker, and
  Narayanan}]{scheffczyk2010reasoning}
Jan Scheffczyk, Collin~F Baker, and Srini Narayanan. 2010.
\newblock Reasoning over natural language text by means of framenet and
  ontologies.
\newblock \emph{Ontology and the lexicon: A natural language processing
  perspective}, pages 53--71.

\bibitem[{Seo et~al.(2016)Seo, Kembhavi, Farhadi, and
  Hajishirzi}]{Seo2016BidirectionalAF}
Min~Joon Seo, Aniruddha Kembhavi, Ali Farhadi, and Hannaneh Hajishirzi. 2016.
\newblock Bidirectional attention flow for machine comprehension.
\newblock \emph{ArXiv}, abs/1611.01603.

\bibitem[{Singh et~al.(2002)Singh, Lin, Mueller, Lim, Perkins, and
  Zhu}]{Singh:2002:OMC:646748.701499}
Push Singh, Thomas Lin, Erik~T. Mueller, Grace Lim, Travell Perkins, and Wan~Li
  Zhu. 2002.
\newblock \href {http://dl.acm.org/citation.cfm?id=646748.701499} {Open mind
  common sense: Knowledge acquisition from the general public}.
\newblock In \emph{On the Move to Meaningful Internet Systems, 2002 -
  DOA/CoopIS/ODBASE 2002 Confederated International Conferences DOA, CoopIS and
  ODBASE 2002}, pages 1223--1237, Berlin, Heidelberg. Springer-Verlag.

\bibitem[{Speer et~al.(2016)Speer, Chin, and Havasi}]{speer2016conceptnet}
Robyn Speer, Joshua Chin, and Catherine Havasi. 2016.
\newblock \href {http://arxiv.org/abs/1612.03975} {Conceptnet 5.5: An open
  multilingual graph of general knowledge}.
\newblock In \emph{AAAI Conference on Artificial Intelligence}.

\bibitem[{Sun et~al.(2019)Sun, Yu, Chen, Yu, Choi, and Cardie}]{TACL1534}
Kai Sun, Dian Yu, Jianshu Chen, Dong Yu, Yejin Choi, and Claire Cardie. 2019.
\newblock \href {https://www.transacl.org/ojs/index.php/tacl/article/view/1534}
  {Dream: A challenge dataset and models for dialogue-based reading
  comprehension}.
\newblock \emph{Transactions of the Association for Computational Linguistics},
  7:217--231.

\bibitem[{Talmor et~al.(2019)Talmor, Herzig, Lourie, and
  Berant}]{talmor-etal-2019-commonsenseqa}
Alon Talmor, Jonathan Herzig, Nicholas Lourie, and Jonathan Berant. 2019.
\newblock \href {https://doi.org/10.18653/v1/N19-1421} {{C}ommonsense{QA}: A
  question answering challenge targeting commonsense knowledge}.
\newblock In \emph{Proceedings of the 2019 Conference of the North {A}merican
  Chapter of the Association for Computational Linguistics: Human Language
  Technologies, Volume 1 (Long and Short Papers)}, pages 4149--4158,
  Minneapolis, Minnesota. Association for Computational Linguistics.

\bibitem[{Tandon et~al.(2018)Tandon, Dalvi, Grus, Yih, Bosselut, and
  Clark}]{tandon-etal-2018-reasoning}
Niket Tandon, Bhavana Dalvi, Joel Grus, Wen-tau Yih, Antoine Bosselut, and
  Peter Clark. 2018.
\newblock \href {https://doi.org/10.18653/v1/D18-1006} {Reasoning about actions
  and state changes by injecting commonsense knowledge}.
\newblock In \emph{Proceedings of the 2018 Conference on Empirical Methods in
  Natural Language Processing}, pages 57--66, Brussels, Belgium. Association
  for Computational Linguistics.

\bibitem[{Wang et~al.(2017)Wang, Yang, Wei, Chang, and
  Zhou}]{wang-etal-2017-gated}
Wenhui Wang, Nan Yang, Furu Wei, Baobao Chang, and Ming Zhou. 2017.
\newblock \href {https://doi.org/10.18653/v1/P17-1018} {Gated self-matching
  networks for reading comprehension and question answering}.
\newblock In \emph{Proceedings of the 55th Annual Meeting of the Association
  for Computational Linguistics (Volume 1: Long Papers)}, pages 189--198,
  Vancouver, Canada. Association for Computational Linguistics.

\bibitem[{Weissenborn et~al.(2018)Weissenborn, Kovcisk'y, and
  Dyer}]{Weissenborn2018DynamicIO}
Dirk Weissenborn, Tom'avs Kovcisk'y, and Chris Dyer. 2018.
\newblock Dynamic integration of background knowledge in neural nlu systems.

\bibitem[{Xiong et~al.(2016)Xiong, Zhong, and Socher}]{Xiong2016DynamicCN}
Caiming Xiong, Victor Zhong, and Richard Socher. 2016.
\newblock Dynamic coattention networks for question answering.
\newblock \emph{ArXiv}, abs/1611.01604.

\bibitem[{Yang et~al.(2019)Yang, Dai, Yang, Carbonell, Salakhutdinov, and
  Le}]{yang2019xlnet}
Zhilin Yang, Zihang Dai, Yiming Yang, Jaime Carbonell, Ruslan Salakhutdinov,
  and Quoc~V. Le. 2019.
\newblock \href {http://arxiv.org/abs/1906.08237} {Xlnet: Generalized
  autoregressive pretraining for language understanding}.
\newblock Cite arxiv:1906.08237Comment: Pretrained models and code are
  available at https://github.com/zihangdai/xlnet.

\bibitem[{Yu et~al.(2018)Yu, Dohan, Luong, Zhao, Chen, and Le}]{46691}
Adams~Wei Yu, David Dohan, Thang Luong, Rui Zhao, Kai Chen, and Quoc Le. 2018.
\newblock \href {https://openreview.net/pdf?id=B14TlG-RW} {Qanet: Combining
  local convolution with global self-attention for reading comprehension}.

\bibitem[{Zellers et~al.(2018{\natexlab{a}})Zellers, Bisk, Farhadi, and
  Choi}]{DBLP:journals/corr/abs-1811-10830}
Rowan Zellers, Yonatan Bisk, Ali Farhadi, and Yejin Choi. 2018{\natexlab{a}}.
\newblock \href {http://arxiv.org/abs/1811.10830} {From recognition to
  cognition: Visual commonsense reasoning}.
\newblock \emph{CoRR}, abs/1811.10830.

\bibitem[{Zellers et~al.(2018{\natexlab{b}})Zellers, Bisk, Schwartz, and
  Choi}]{zellers-etal-2018-swag}
Rowan Zellers, Yonatan Bisk, Roy Schwartz, and Yejin Choi. 2018{\natexlab{b}}.
\newblock \href {https://doi.org/10.18653/v1/D18-1009} {{SWAG}: A large-scale
  adversarial dataset for grounded commonsense inference}.
\newblock In \emph{Proceedings of the 2018 Conference on Empirical Methods in
  Natural Language Processing}, pages 93--104, Brussels, Belgium. Association
  for Computational Linguistics.

\bibitem[{Zellers et~al.(2019)Zellers, Holtzman, Bisk, Farhadi, and
  Choi}]{zellers-etal-2019-hellaswag}
Rowan Zellers, Ari Holtzman, Yonatan Bisk, Ali Farhadi, and Yejin Choi. 2019.
\newblock \href {https://www.aclweb.org/anthology/P19-1472} {{H}ella{S}wag: Can
  a machine really finish your sentence?}
\newblock In \emph{Proceedings of the 57th Annual Meeting of the Association
  for Computational Linguistics}, pages 4791--4800, Florence, Italy.
  Association for Computational Linguistics.

\bibitem[{Zhang et~al.(2018)Zhang, Liu, Liu, Gao, Duh, and
  Durme}]{DBLP:journals/corr/abs-1810-12885}
Sheng Zhang, Xiaodong Liu, Jingjing Liu, Jianfeng Gao, Kevin Duh, and
  Benjamin~Van Durme. 2018.
\newblock \href {http://arxiv.org/abs/1810.12885} {Record: Bridging the gap
  between human and machine commonsense reading comprehension}.
\newblock \emph{CoRR}, abs/1810.12885.

\bibitem[{Zhong et~al.(2018)Zhong, Tang, Duan, Zhou, Wang, and
  Yin}]{DBLP:journals/corr/abs-1809-03568}
Wanjun Zhong, Duyu Tang, Nan Duan, Ming Zhou, Jiahai Wang, and Jian Yin. 2018.
\newblock \href {http://arxiv.org/abs/1809.03568} {Improving question answering
  by commonsense-based pre-training}.
\newblock \emph{CoRR}, abs/1809.03568.

\end{thebibliography}
\bibliographystyle{acl_natbib}

\appendix

\end{document}